\pdfoutput=1

\documentclass[11pt]{article}

\usepackage[]{acl}

\usepackage{times}
\usepackage{latexsym}
\usepackage{graphicx}

\usepackage[T1]{fontenc}

\usepackage[utf8]{inputenc}

\usepackage{microtype}

\usepackage{inconsolata}

\usepackage{amsmath,amsfonts,bm}

\def\eqref#1{equation~\ref{#1}}

\def\1{\bm{1}}

\DeclareMathAlphabet{\mathsfit}{\encodingdefault}{\sfdefault}{m}{sl}
\SetMathAlphabet{\mathsfit}{bold}{\encodingdefault}{\sfdefault}{bx}{n}

\DeclareMathOperator*{\argmax}{arg\,max}

\usepackage{hyperref}
\usepackage{xspace}
\usepackage{amsmath}
\usepackage{amssymb}
\usepackage{booktabs}
\usepackage{float}
\usepackage{amsmath}
\usepackage{pifont}%

\newcommand{\method}{\textsc{Byokg}\xspace}

\newcommand{\answerset}{\ensuremath{\mathcal{A}}}

\newcommand{\exploration}{\ensuremath{\mathcal{X}}}
\newcommand{\explorationprograms}{\ensuremath{\mathcal{P}^\mathcal{X}}}
\newcommand{\kg}{\ensuremath{\mathcal{K}}}
\newcommand{\entities}{\ensuremath{\mathcal{E}}}
\newcommand{\relations}{\ensuremath{\mathcal{R}}}
\newcommand{\literals}{\ensuremath{\mathcal{L}}}
\newcommand{\classes}{\ensuremath{\mathcal{C}}}
\newcommand{\query}{\ensuremath{q}}
\newcommand{\program}{\ensuremath{p}}
\newcommand{\ours}{\scriptsize{(\textsc{ours})}}
\newcommand{\sota}{\scriptsize{(\textsc{sota})}}
\newcommand{\pangu}{Pangu} %
\newcommand{\zeroshot}{Zero-shot} %
\newcommand*{\argtopk}{\mathop{\mathrm{arg\,topk}}}
\newcommand{\LLM}{\mathtt{LLM}}
\newcommand{\fulltrainset}{\ensuremath{\mathcal{T}}}
\newcommand{\trainset}{\ensuremath{\mathcal{T}_{10k}}}
\newcommand{\trainsetk}{\ensuremath{\mathcal{T}_{1k}}}

\newcommand{\ulb}[1]{\textbf{\ul{#1}}}

\newcount\Comments  %
\definecolor{darkgreen}{rgb}{0,0.5,0}
\definecolor{darkred}{rgb}{0.7,0,0}
\definecolor{teal}{rgb}{0.1,0.6,0.7}
\definecolor{blue}{rgb}{0.0,0.1,0.9}
\definecolor{orange}{rgb}{1.,0.7,0.0}
\definecolor{lightblue}{rgb}{0.70, 0.80, 0.89}
\definecolor{violet}{rgb}{0.50, 0.16, 0.88}
\newcommand{\kibitz}[2]{\ifnum\Comments=1{{\textcolor{#1}{\textsf{\footnotesize [#2]}}}}\fi}

\makeatletter
\def\adl@drawiv#1#2#3{%
        \hskip.5\tabcolsep
        \xleaders#3{#2.5\@tempdimb #1{1}#2.5\@tempdimb}%
                #2\z@ plus1fil minus1fil\relax
        \hskip.5\tabcolsep}
\newcommand{\cdashlinelr}[1]{%
  \noalign{\vskip\aboverulesep
           \global\let\@dashdrawstore\adl@draw
           \global\let\adl@draw\adl@drawiv}
  \cdashline{#1}
  \noalign{\global\let\adl@draw\@dashdrawstore
           \vskip\belowrulesep}}
\makeatother

\usepackage{multirow}
\usepackage{arydshln}
\usepackage{changepage}
\usepackage{soul}
\usepackage{listings}
\newfloat{lstfloat}{htbp}{lop}
\floatname{lstfloat}{Prompt}

\title{\textsc{Bring Your Own KG}: \\Self-Supervised Program Synthesis for Zero-Shot KGQA}

\author{Dhruv Agarwal$^{1,}$\thanks{~Work done during an internship at Amazon. ~~~~~~~~~~~~~~~~$^{\dagger}$~Corresponding authors. Our code is available at \url{https://github.com/amazon-science/BYOKG-NAACL24}.}\ , Rajarshi Das$^{2,\dagger}$, Sopan Khosla$^{2,\dagger}$, Rashmi Gangadharaiah$^2$ 
\\ $^1$University of Massachusetts Amherst, $^2$AWS AI Labs \\ \texttt{dagarwal@cs.umass.edu}, \ \texttt{\{dasrajar,sopankh,rgangad\}@amazon.com} }

\begin{document}
\maketitle
\begin{abstract}

We present \method, a universal question-answering (QA) system that can operate on \emph{any} knowledge graph (KG), requires no human-annotated training data, and can be ready to use within a day---attributes that are out-of-scope for current KGQA systems.
\method draws inspiration from the remarkable ability of humans to comprehend information present in an unseen KG through exploration---starting at random nodes, inspecting the labels of adjacent nodes and edges, and combining them with their prior world knowledge.
Exploration in \method
leverages an LLM-backed symbolic agent that generates a diverse set of query-program exemplars, which are then used to ground a retrieval-augmented reasoning procedure to synthesize
programs for arbitrary questions.
\method is effective over both small- and large-scale graphs, showing dramatic gains in zero-shot QA accuracy of 27.89 and 59.88 F1 on GrailQA and MetaQA, respectively.
We further find that performance of \method reliably improves with continued exploration as well as improvements in the base LLM, notably outperforming a state-of-the-art \emph{fine-tuned} model by 7.08 F1 on a sub-sampled zero-shot split of GrailQA. 
Lastly, we verify our universality claim by evaluating \method on a domain-specific materials science KG and show that it improves zero-shot performance by 46.33 F1.

\end{abstract}

\begin{figure*}
    \centering
    \includegraphics[width=1.0\textwidth]{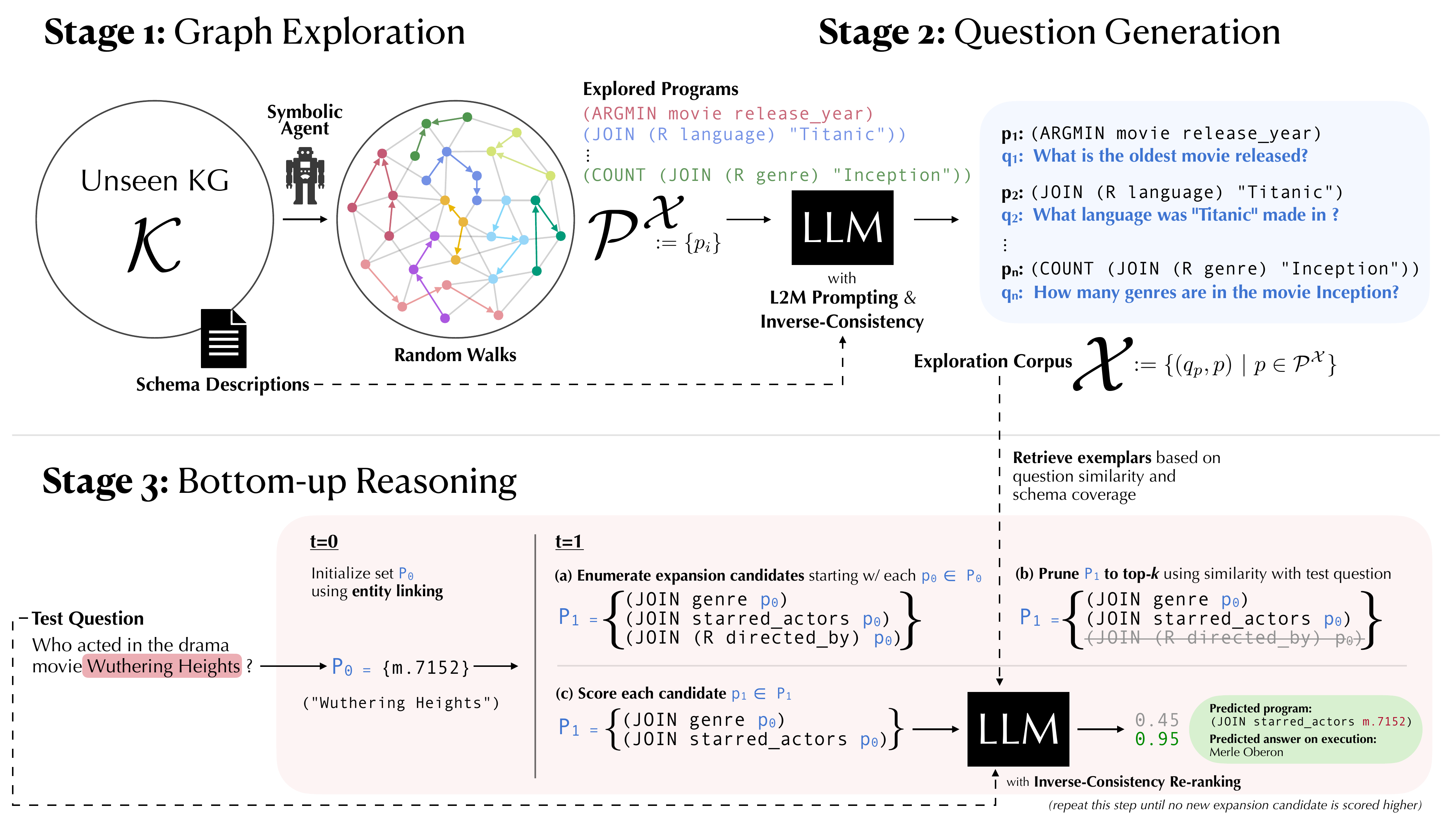}
    \caption{\textbf{Overview.} Given a new KG, a symbolic graph explorer 
    generates diverse programs. 
    Next, an LLM generates 
    questions 
    for the 
    programs 
    using descriptions of schema items, which are then stored in 
    an exploration corpus. 
    This process is done once for a KG. 
    To answer a given question, \method adopts a grounded reasoning approach that iteratively synthesizes the correct program using retrieved 
    exemplars from the exploration corpus.}
    \label{fig:overview}
\end{figure*}

\section{Introduction}
\label{sec:intro}
The ability to query structured data stores such as knowledge graphs (KGQA) via natural language is crucial for making the information within them accessible ~\citep{liang2016learning,mythesis}.
However, most prior works that aim to create such interfaces assume the availability of \textit{some} training data (query-program pairs)~\citep{talmor2018web,keysers2020measuring,gu2021beyond,Dutt2023,kgcomplexlm}, which, in practice, might be unrealistic. 
For example, in scientific domains such as materials science and clinical decision-making, training data may be completely unavailable due to high collection costs or stringent privacy regulations~\citep{sima2022bio}. 
Further, even when training data is available, models trained on one dataset may not generalize o.o.d. to other datasets of the \emph{same} KG~\citep{khosla-etal-2023-exploring}.

In this work, we, therefore, set out to answer the following question---\ul{can we develop a universal QA system 
that is ready for use 
with \emph{any} KG,
within a reasonable amount of time (e.g., 24 hours),
and without \textit{any} training data?}
To achieve this, a model must efficiently and accurately learn to reason over a KG with no prior knowledge of the query distribution
or the KG semantics.

\textbf{\method} takes inspiration from the human tendency to be curious---seeking challenges and developing knowledge even in the absence of well-defined rewards~\citep{oudeyer2016intrinsic,di2017emerging}. 
Given a new KG, a human practitioner begins familiarizing themselves with the graph by inspecting random nodes and analyzing the various properties\footnote{For e.g., \url{https://prop-explorer.toolforge.org/}.} found in the node neighborhoods.
As this process continues (crucially, without a task-specific information need in mind), the practitioner develops an intuition for the set of questions that can be answered with the information present in the KG.

To mechanize this human tendency, \method consists of an exploration module, which combines random walks over the KG nodes with a set of graph operations (e.g. \texttt{COUNT}, \texttt{ARGMAX}, \texttt{>=}, etc.) to produce programs of varying degrees of complexity (\underline{\textsc{Stage 1}}; fig.~\ref{fig:overview}).
Our explorer is symbolic in nature and has the goal of maximizing diversity within the generated programs, akin to curiosity-driven human learning \citep{ryan2000intrinsic}.

After sampling a diverse set of programs, \method leverages the strong generalization ability of large language models (LLMs)~\citep{brown2020language,wei2022emergent,touvron2023llama} to generate questions 
for each program (\underline{\textsc{Stage 2}}).
However, we find that LLM outputs are often semantically inaccurate with respect to the program,
particularly in the zero-shot setting. 
To improve LLM generation, we, thus,
develop a novel inverse-consistency re-ranking method,
which computes scores for generated queries based on the likelihood of the query \textit{re-generating} the program. We also incorporate least-to-most (L2M) prompting \citep{zhou2023leasttomost} to improve generation for multi-hop programs.
Empirically, we find that both 
techniques
greatly improve the accuracy of question generation and are essential in allowing us to operate within our unsupervised setting.

Finally, \method uses the explored query-program pairs to perform reasoning in order to answer user queries (\underline{\textsc{Stage 3}}).
With the 
motivation of designing a QA system that can work on any KG, we opt for a semi-parametric
approach instead of KG-specific fine-tuning. In particular, we build upon Pangu \citep{gu-etal-2023-dont}, an 
LLM-based discriminative procedure that iteratively synthesizes the predicted program guided by 
retrieved exemplars 
from the training data. 
We introduce several modifications,
including a pruning step, which dramatically reduces runtime (by 88\%) as well as increases accuracy.

In summary, our contributions are as follows---\ulb{(a)} we introduce \textbf{\method}, a method that allows practitioners to ``bring their own KG'' with \textit{no} training data 
and have a natural language query interface ready within a day. 
\ulb{(b)} Inspired by intrinsic motivation,
we develop an LLM-backed exploration module, which explores the KG to gather query-program exemplars. 
We demonstrate that ICL-based models that use our exploration perform competitively with models that use annotated training data.
\ulb{(c)} We show that our proposed inverse-consistency re-ranking and L2M prompting greatly improve the quality of zero-shot generation.
\ulb{(d)} We demonstrate that \method is effective on both small- (MoviesKG; ${10}^5$ edges) %
and large-scale KGs (Freebase; 3$\times{10}^9$ edges). %
On GrailQA and MetaQA, \method provides dramatic improvements of 27.89 and 59.88 F1, respectively, over a zero-shot baseline. \ulb{(e)} We show that \method scales with model size and even \textit{outperforms} a state-of-the-art fine-tuned model on zero-shot queries by 7.08 F1 on GrailQA using a larger LM (GPT-3.5).
\ulb{(f)} Finally, we demonstrate that \method is able to operate in arbitrary domains without training data, showing a strong 46.33 F1 gain using a materials science KG.

\section{Task Definition}
\vspace{-2mm}
\paragraph{KGQA.} A knowledge graph $\mathcal{K}$ is a set of triples, or facts, of the form $\entities \times \relations \times (\entities~\cup~\literals~\cup~\classes)$, where \entities, \relations, \literals, and \classes~denote entities, binary relations, literals, and classes (entity types), respectively. KGQA is then defined as the task of finding a set of answers \answerset~over graph \kg~for a natural language question \query. In program synthesis, the task is evaluated as mapping \query~to a program $\program_\query$ (e.g. SPARQL or s-expression \citep{su-etal-2016-generating}), which can deterministically be executed using a query engine to generate the answer set, i.e. $\mathtt{eval}^\kg(\program_\query)=\answerset_q$.

\paragraph{Unsupervised KGQA.} We define \textit{unsupervised} KGQA as a zero-shot setting where no
query supervision over the target distribution is available\footnote{This is a stronger generalization requirement than prior work \citep{gu2021beyond}, where queries with even a single schema item unseen at training are considered zero-shot.}.
Unsupervised KGQA jointly addresses multiple dimensions of generalization---linguistic variability~\citep{khosla-etal-2023-exploring}, query complexity~\citep{keysers2020measuring,gu2021beyond, kgcomplexlm}, domain transfer~\citep{gu2021beyond,baek2023knowledge}, and schema generalization~\citep{das-etal-2021-case,badenes2023muheqa}---each of which has individually been shown to pose challenges to current QA systems.

\section{Method}
\label{sec:method}
\method consists of three stages---\textbf{graph exploration} (\S{\ref{subsec:graph_exploration}}), \textbf{query generation} (\S{\ref{subsec:query_generation}}), and \textbf{reasoning} (\S{\ref{subsec:reasoning}}). First, our method explores the KG to enumerate a diverse set of executable programs. Next, each explored program is converted into a natural language question by prompting an LLM with schema descriptions of the relations and classes in the program. Finally, 
\method leverages its acquired knowledge from exploration to ground a bottom-up inference procedure
to iteratively generate the final program.

\subsection{Symbolic Graph Exploration}
\label{subsec:graph_exploration}
The goal of graph exploration is to enumerate 
possible programs 
that may be queried at test time. 
However, exhaustive enumeration is often impractical with real-world KGs due to limited compute and time budgets. 
Instead, we construct a set of \textbf{explored programs} \explorationprograms~ that provides approximate coverage of query patterns supported by the KG.
\method uses a symbolic, graph-based \citep{su-etal-2016-generating} random walk procedure to enumerate a diverse set of executable programs. 

Concretely, a symbolic agent begins exploration by initializing a sub-program $\program_0$ at $t=0$ with a class $c_0 \sim \classes$. Next, the agent determines $S_{\program_0} := \{s~\big|~s \in \mathcal{R}~\cup~\mathcal{C} : \mathtt{reachable}(\program_0, s)\}$, the set of schema items reachable from $\program_0$. The agent then picks an item $s_0 \sim S_{\program_0}$ to extend the sub-program into $\program_1$. 
This process is repeated until 
the desired complexity of the program (i.e. relation count) is satisfied. The agent then, optionally, samples a program function $f \sim \mathcal{F}$ to apply over $\program_t$, where $\mathcal{F}$ contains operators such as \texttt{COUNT}, comparatives, and superlatives. To encourage diversity, we discard $\program_t$ and repeat the process if \explorationprograms~already contains $\program_t$\footnote{We set the max. number of programs per pattern to 5.}. Finally, we ground the classes appearing in $p_t$ randomly by sampling from $\{e~\big|~e \in \entities : \mathtt{eval}^\kg(\program^e_t) \neq \emptyset\}$, the set of entities that lead to non-empty answer sets on program execution. The grounded $\program^e_t$ is then added to \explorationprograms.

\subsection{Natural Language Query Generation}
\label{subsec:query_generation}
For each $\program \in \explorationprograms$, we next generate a natural language question $\query_\program$ to build an \textbf{exploration corpus} $\exploration:=\{(\query_\program,\program)~\big|~\program \in \explorationprograms\}$ of query-program pairs. 
To generate questions, we prompt an LLM with instructions and textual descriptions of schema items relevant to each program (see~\ref{app: schema_supervision}).
Generating accurate output without in-context exemplars, however, is challenging.
To elicit reliable zero-shot generation, we, therefore, utilize two techniques---(1) least-to-most prompting \citep{zhou2023leasttomost}, which generates outputs for complex programs in a step-by-step manner, and (2) a novel \textit{inverse-consistency} method to re-rank LLM generations by scoring the inverse task of program generation.

\subsubsection{Least-to-Most Prompting}
\label{subsubsec:least_to_most_prompting}
Several prior works \citep{jung-etal-2022-maieutic, zhou2023leasttomost, drozdov2023compositional} have tackled complex generative tasks by providing intermediate supervision via iteratively prompting the model with its own generations as additional context. 
Using these observations, we implement a least-to-most (L2M) prompting strategy that first decomposes $\program$ into simpler sub-programs $(p^1, p^2, \ldots, p^n)$ of increasing complexity using bottom-up parsing. We then generate a question $\query_{\program^i}$ for each sub-program, appending each $(\program^j, \query_{\program^j})$ for $j < i$ as additional demonstrations in the prompt (see~\ref{app: lm_prompts_qgen_l2m}).\footnote{Query decomposition with s-expressions is straightforward---starting from the inner-most clause, the next sub-program is generated by simply including all the terms within the next parenthetic level.} In \ref{subsubsec:analysis-l2m}, we show that L2M is crucial in unlocking deliberate, ``System 2'' reasoning~\citep{kahneman2011thinking} for complex queries in the zero-shot setting.

\subsubsection{Inverse-Consistency Re-ranking}
\label{subsubsec:inv_consistency_qgen}
We observe that even when LLMs \textit{can} produce the right answer within a top-$k$ set of generations (e.g., from beam search), they do not always rank the correct answer as the top prediction, particularly with smaller models and in the unsupervised setting, rendering their use infeasible (see~\ref{app: qualitative_examples_inv_generation}).
To tackle this, we introduce a re-ranking mechanism that scores output sequences from an LM using the likelihood of an \textit{inverse} task, i.e. how likely the \textit{input} sequence is given the output. 

Concretely, consider a generative task $T := y\ |\ I, D, x$, where $x$ is a sequence of query tokens, $y$ is the target sequence of tokens to be predicted by a decoding algorithm, 
$I$ is the textual instruction for the task, and $D$ is the set of in-context demonstrations ($D=\emptyset$ in the unsupervised setting). The prediction $y_{\mathrm{pred}}$ for $T$ is the top-ranked sequence from a list of candidates $\textbf{y}_{\mathrm{cands}}$ generated by the decoding algorithm measured using length-normalized log-probability scores, i.e. $y_{\mathrm{pred}} := \argmax_{y \in \textbf{y}_{\mathrm{cands}}} \log\mathrm{Pr}(y\ |\ I, D, x)~/~|y|$.
To re-rank $\textbf{y}_\mathrm{cands}$, we now construct the following \textit{inverse} task: 
$$T^{-1} := x\ |\ I^{-1}, D^{-1}, y,$$
i.e. the task of predicting the query sequence $x$ given an output sequence $y$ from $T$, along with a new instruction $I^{-1}$ for the inverse task and, optionally, an inverted demonstration set $D^{-1}$. For e.g., for the task of query generation, the inverse task is program synthesis. The new prediction is then given by
$$y_{\mathrm{pred}} := \argmax_{y \in \textbf{y}_{\mathrm{cands}}} \log\mathrm{Pr}(x\ |\ I^{-1}, D^{-1}, y)~/~|x|.$$
Scoring $T^{-1}$ for a single $y$ requires only one forward pass to get the next-token logit distribution at each position, allowing efficient computation of the log-probability score of the fixed-sequence $x$ given $y$. Scores over the entire set $\textbf{y}_\mathrm{cands}$ can simply be computed using a batched forward pass. 
Inverse-consistency indeed improves generation accuracy (\ref{subsubsec:analysis-inv_consistency_q_reranking}) and enables \method to use smaller models to scale exploration. We also note the close relation with PMI-scoring \citep{holtzman2021surface}, but observe differing behavior in practice (see~\ref{app: inv_pmi_comparison}).

\subsection{Bottom-up Reasoning}
\label{subsec:reasoning}
With a corpus of query-program pairs in place, we now require a method to synthesize programs given natural language queries at test time. To use a single model with \textit{any} KG, a key desiderata is to avoid KG-specific parameter tuning \citep{khosla-etal-2023-exploring}. 
We, therefore, use an ICL approach using demonstrations from the exploration corpus within an enumerate-and-rank procedure.
We adapt the method in \citet{gu-etal-2023-dont} with modifications that provide speed and accuracy gains to allow \method to operate well in the unsupervised setting.

Concretely, given a test question $\query_{\mathrm{test}}$,
\method first instantiates a set of candidate sub-programs $P_0$ at $t=0$ with all the topic entities, classes, and literals found in the question, extracted using off-the-shelf linkers \citep{li-etal-2020-efficient, agarwal-etal-2022-entity}. In each subsequent timestep $t$, the reasoner determines which sub-programs from the previous step should further be extended. To do this, we use an LLM to compute\footnote{LLM scoring tends to prefer candidates with repeated relations.
We, thus, penalize the final score based on the count of repeated relations. We do not add this penalty on MoviesKG due to the formulaic nature of the evaluation set.} the likelihood of each sub-program being the parse for $\query_\mathrm{test}$ conditioned on retrieved demonstrations $D_\mathrm{test}$ from exploration, and retain the top-$k$ candidates
$$P_{t-1} := \argtopk_{\program^i_{t-1} \in P_{t-1}}\LLM(\program^i_{t-1}, \query_\mathrm{test}, D_\mathrm{test}).$$
We additionally define
$$P_\mathrm{best} := \argtopk_{\program \in P_\mathrm{best}\cup P_{t-1}} \LLM(\program, \query_\mathrm{test}, D_\mathrm{test})$$
as the best set of candidates across timesteps.
After scoring, the reasoner extends each $\program^i_{t-1} \in P_{t-1}$ using an extensible set of program expansion heuristics \citep{gu-su-2022-arcaneqa} to construct the candidate set for the next timestep,
$$P_t := \{\mathtt{extend}(\program^i_{t-1}, S_{\program^i_{t-1}}, P_\mathrm{best} )~\Big|~\program^i_{t-1} \in P_{t-1} \},$$ 
where $S_{\program^i_{t-1}}$ is the set of schema items reachable from $\program^i_{t-1}$ and $P_\mathrm{best}$ is the set of best-$k$ candidates so far. The process terminates when no new sub-program is added to $P_\mathrm{best}$, at which point we output the prediction $\program_\mathrm{pred} := \argmax_{\program \in P_\mathrm{best}}\LLM(\program, \query_\mathrm{test},D_\mathrm{test})$.

\paragraph{ICL from exploration.} To make predictions using an LLM, \method takes a few-shot prompting approach to score candidate sub-programs conditioned on reasoning patterns for similar questions seen during exploration. A typical approach is to retrieve the $k$-most similar exemplars from $\exploration$ using the cosine similarity of exploration queries with the test query as measured using a sentence embedding model \citep{reimers-gurevych-2019-sentence}. Following prior work \citep{thai2023machine}, we additionally anonymize topic entities mentioned within questions to retrieve similar program \textit{patterns} instead of similar topic entities.
For instance, the question \textit{``How many trophies has \underline{Manchester United} won?''} would be anonymized to \textit{``How many trophies has \underline{sports.team} won?''}.

\paragraph{Candidate pruning.} Scoring candidates can entail arbitrary latency depending on the number of candidates to score, making reasoning impractically slow when the candidate set $P_t$ to be scored is very large (Table \ref{tab:reasoning_pruning}).
We, therefore, introduce a candidate pruning step that restricts the size of the candidate set to at most 10 at each step of reasoning based on the similarity of anonymized candidate programs with the anonymized natural language test question using the sentence embedding model from retrieval. To keep our setup KG-agnostic, we do not fine-tune this model. As shown in \ref{subsubsec:analysis-cand_pruning}, we find that not only does pruning improve efficiency, but it also results in more accurate reasoning.

\paragraph{Inverse-consistency for candidate re-ranking.}
When schema items are \textit{completely} unseen during exploration, we find that LLM scoring erroneously assigns high scores to irrelevant candidates that may resemble the retrieved exemplars (see~\ref{app: qualitative_examples_inv_reasoning}).

To address this problem, we re-use inverse-consistency (\S\ref{subsubsec:inv_consistency_qgen}) to \textit{re-rank} the final candidate set $P_\mathrm{best}$. Concretely, we construct the inverse task, denoted by $\LLM^{-1}(\cdot,\cdot)$, to be one of zero-shot question generation. 
To make predictions, we use a weighted combination of the original and inverse scores using weight $\alpha$\footnote{We do not tune $\alpha$, in keeping with our setting of not assuming a dev set, and set its value to $0.5$ in all experiments.}, resulting in
\begin{align*}
    \mathtt{rerank}(\program, \query, D) &:= \alpha \LLM(\program,\query, D)~+ \setminus\\
    & \hspace{1cm}(1-\alpha) \LLM^{-1}(\program,\query),
\end{align*}
which leads to the final prediction
$$\program_\mathrm{pred} := \argmax_{\program \in P_\mathrm{best}}~\mathtt{rerank}(\program, \query_\mathrm{test}, D_\mathrm{test}).$$

\begin{table*}[htb]
    \small
    \centering
    \begin{tabular}{l l l c  c  c  c }
    \toprule
        & \textbf{Method} & \textbf{Model} & \bf Overall & \bf I.I.D. & \bf Compositional & \bf Zero-shot \\
        \midrule
         \multirow{2}{*}{\textit{Supervised}} & \pangu-FT~\sota & T5-3B & \bf 81.7 & \bf 88.8 & \bf 81.5 & \bf 78.5 \\
         \cdashlinelr{2-7}
         \textit{(w/ train set)} & \pangu-ICL~+~\trainsetk & Codex & 65.0 & 73.7 & 64.9 & 61.1 \\
         &  \pangu-ICL$^{\dagger}$~+~\trainset & MPT-7B & 44.67 & 58.15 & 40.90 & 40.15 \\
         & \method~+~\trainset & MPT-7B & 46.61 & 58.29 & 45.14 & 41.89 \\
        \midrule
         \ul{\textbf{Unsupervised}} & \zeroshot & MPT-7B & 18.58 & 19.13 & 16.34 & 19.33 \\
         & \pangu-ICL$^{\dagger}$~+ \exploration & MPT-7B & 42.44 ($\Delta$+23.86) & 45.08 & 38.79 & 42.85 \\
         & \textbf{\method} + \exploration~\ours & MPT-7B & \textbf{46.47} ($\Delta$+27.89) & \bf 48.91 & \bf 43.22 & \bf 46.80 \\
    \bottomrule
    \end{tabular}
    \caption{\textbf{KGQA Results on GrailQA.} F1-scores for \method in the unsupervised setting on the GrailQA test set compared to a zero-shot baseline and \pangu. For reference, we also report performance with models that use training data---ICL with randomly sampled training exemplars (\trainsetk~and \trainset) as well as a state-of-the-art fine-tuned model. We find that \method~+~\exploration~improves zero-shot performance by 2.5x (nearly matching the performance of its supervised counterpart). \method also demonstrates stable performance across generalization splits ($\sigma=$~2.35), whereas supervised methods ($\sigma=$~7.09) show drops in performance on the compositional and zero-shot splits. ($\dagger$ indicates our re-implementaton)
    }
    \label{tab:kgqa_grailqa}
\end{table*}

\begin{table}[htb]
    \small
    \centering
    \begin{adjustwidth}{-0.125cm}{}
    \begin{tabular}{l c c cc cc cc}
    \toprule
        \bf Model & \bf Overall & \bf I.I.D. & \bf Comp. & \bf Z-shot \\
        \midrule
         \pangu-FT & \bf 81.68 & \bf 92.81 & \bf 79.97 & 73.91 \\
         \pangu & \hspace{-1.75cm}-Codex \hspace{0.7cm} 65.0 & 73.7 & 64.9 & 61.1 \\
         \cdashlinelr{1-5}
         \textbf{\method} & \hspace{-2.2cm} +~\exploration~\ours &&& \\
         \hspace{0.1cm} MPT-7B & 66.79 & 70.40 & 61.35 & 69.08 \\ 
         \hspace{0.1cm} MPT-30B & 69.58 ($\Delta$+2.79) & 73.10 & 65.14 & 70.95 \\ 
         \hspace{0.1cm} GPT-3.5 & 75.16 ($\Delta$+8.37) & 73.89 & 70.33 & \bf 80.99 \\ 
    \bottomrule
    \end{tabular}
    \end{adjustwidth}
    \caption{\textbf{\method Accuracy v/s Model Scale.} F1-scores for \method~+~\exploration~using 300 randomly sampled questions
    from the GrailQA dev set.
    (a) \method shows gains in accuracy with improvements in the underlying LLM. (b) \method with GPT-3.5 shows stable performance across generalization splits (unlike \pangu~with training data). (c) \method \textit{outperforms} \pangu-FT on the zero-shot split by 7.08 points. ($^*$\textit{Note:} \pangu-Codex test set results are included only to provide an estimate of ICL performance with a similar model.)}
    \label{tab:kgqa_grailqa_gpt}
\end{table}

\begin{table*}[htb]
    \small
    \centering
    \begin{adjustwidth}{-0.4cm}{}
    \begin{tabular}{l l c c c c  c c  c c}
    \toprule
        & & \multicolumn{2}{c}{\bf~~~~~~~~Overall} & \multicolumn{2}{c}{\bf 1-hop} & \multicolumn{2}{c}{\bf 2-hop} & \multicolumn{2}{c}{\bf 3-hop}\\
        \cmidrule(lr){3-10}
        & \textbf{Method} & \textbf{F1} & \textbf{Hits@1} & \textbf{F1} & \textbf{Hits@1} & \textbf{F1} & \textbf{Hits@1} & \textbf{F1} & \textbf{Hits@1} \\
        \midrule
         \multirow{2}{*}{\textit{Supervised}} & NSM-FT~\sota & - & \bf 98.82 & - & 97.1 & - & \bf 99.9 & - & \bf 98.9 \\
         \cdashlinelr{2-10}
         \multirow{1}{*}{\textit{(w/ train set)}} & \pangu-ICL$^{\dagger}$~+~\trainset & \bf 85.61 & 92.38 & 97.88 & \bf 98.80 & \bf 93.43 & 94.21 & \bf 69.82 & 86.01 \\
         & \method~+~\trainset & 82.10 & 87.31 & \bf 97.95 & 98.27 & 90.24 & 90.76 & 62.57 & 76.08 \\
        \midrule
         \ul{\textbf{Unsupervised}} & \zeroshot & 15.43 & 25.11 & 34.07 & 41.67 & 8.10 & 11.42 & 10.09 & 27.84 \\
         & \pangu-ICL$^{\dagger}$~+ \exploration & 54.68 ($\Delta$+39.25) & 64.87 & 59.32 & 63.40 & 62.67 & 66.74 & 44.60 & 63.96 \\
         & \textbf{\method} + \exploration~\ours & \textbf{75.31} ($\Delta$+59.88) & \textbf{83.01} & \textbf{94.83} & \textbf{95.25} & \textbf{80.28} & \textbf{81.85} & \textbf{56.54} & \textbf{75.69} \\
    \bottomrule
    \end{tabular}
    \end{adjustwidth}
    \caption{\textbf{KGQA Results on MetaQA.} F1-scores for \method in the unsupervised setting on the MetaQA test set compared to a zero-shot baseline and \pangu. For reference, we also report supervised ICL baselines with 10K randomly sampled training examples (\trainset) and NSM, a state-of-the-art fine-tuned LSTM. Exploration (\exploration) improves zero-shot F1 performance by 3.5x using \pangu~and 4.9x using \method. Further, \method~+~\exploration~closes the gap with the best-performing supervised baseline to within only 10.3 F1. ($\dagger$ indicates our re-implementaton; all ICL methods are evaluated using MPT-7B.)}
    \label{tab:kgqa_metaqa}
\end{table*}

\section{Experiments}
\label{sec:experiments}

\subsection{Graphs and Datasets}
\label{subsec:datasets_graphs}
For our larger-scale experiments, we use \textbf{Freebase}~\citep{bollacker2008freebase} and evaluate QA performance on the \textbf{GrailQA}~\citep{gu2021beyond} dataset. For smaller, domain-specific evaluation, we use \textbf{MoviesKG}~\citep{miller2016key} and the \textbf{MetaQA}~\citep{zhang2017variational} dataset. Note that in the unsupervised setting, all datasets are o.o.d..\footnote{See Appendix \ref{app: dataset_stats} for details on the datasets and KGs.}

\subsection{Evaluation Metrics}
\label{subsec:eval_metrics}
Our primary metric is the \textbf{F1-score} between the predicted and reference answer sets. Several prior works (on MetaQA), however, only provide ranked entities. To compare, we report \textbf{Hits@1}, assigning rank 1 to each answer in our prediction set.

\subsection{Models}
\label{subsec:models}
We use \textbf{MPT-Instruct}~\citep{MosaicML2023Introducing} (7B) for our main experiments. To demonstrate the scaling behavior of \method, we additionally use \textbf{MPT-30B} as well as \textbf{GPT-3.5}~\citep{brown2020language} with the \textit{text-davinci-003} variant\footnote{Of the available variants, only \textit{text-davinci}, \textit{text-curie}, and \textit{text-babbage} are compatible with \method since we require access to log-probabilities to score sequences.}.

\subsection{Experimental Settings}
\label{subsec:baselines}
\subsubsection{Unsupervised}
Our main experimental setting evaluates models with no access to \emph{any} query supervision. 

\textbf{Zero-shot} represents our bottom-up reasoning procedure from \S{\ref{subsec:reasoning}} but without any in-context demonstrations to score sub-programs at each step. 

\textbf{ICL~+~Exploration} represents our proposed \method method. In this setting, in-context demonstrations are retrieved from the exploration corpus \exploration, which we limit to 10K programs based on our time and compute budget. We also include in this setting results with \pangu-ICL~\citep{gu-etal-2023-dont}, the few-shot variant of a KGQA method closely related to the bottom-up reasoning procedure of \method.

\subsubsection{Supervised}
To situate our evaluations in the unsupervised setting, we also include a comparison with methods that have access to curated training data.

\textbf{ICL~+~Train Set} is the setting where both \method and \pangu~retrieve demonstrations from a randomly sampled subset of 10K training exemplars \trainset. On GrailQA, we also report published \pangu-ICL (1000-shot) results with OpenAI Codex~\citep{chen2021evaluating}\footnote{LLM for instruction-following on code (now deprecated).}.

\textbf{Fine-tuned} includes \pangu-FT, a fine-tuned T5-3B \citep{2020t5} variant of \pangu~trained using the full train set of 44K exemplars on GrailQA, and is currently the state-of-the-art (without ensembling). On MetaQA, we include NSM-FT \citep{he2021improving}, a fine-tuned method trained using teacher-student networks over 329K training exemplars. Although these models comprise dataset-specific parameters, we include them to provide an estimate of an upper-bound\footnote{No strict bound exists for unsupervised performance to be lower than supervised. See Fig.~\ref{fig:budget_kgqa} for scaling trends.}.

\section{Results}
\label{subsec:results}

\paragraph{Exploration leads to substantial gains in the unsupervised setting.} On both GrailQA with the Freebase KG (Table~\ref{tab:kgqa_grailqa}) and MetaQA with the MoviesKG (Table~\ref{tab:kgqa_metaqa}), we find that unsupervised exploration leads to dramatic gains over the zero-shot baseline. Specifically, our proposed \method~+~\exploration~results in large 27.89 F1 (2.5x) and 59.88 F1 (4.9x) improvements on GrailQA and MetaQA, respectively.

\paragraph{\method exhibits better compositional generalization than \pangu.} On GrailQA, \method outperforms \pangu~by 4.03 F1 (Table~\ref{tab:kgqa_grailqa}) and on MetaQA by a large 20.63 F1 (Table~\ref{tab:kgqa_metaqa}) when evaluated with our exploration corpus.   Note that exploration provides only partial coverage over evaluation queries (as shown in Table~\ref{tab:exploration_coverage}). Therefore, models must compositionally assemble sub-expressions from relevant exemplars to make predictions. For instance, on MetaQA, we find that training data provides perfect test pattern coverage, which translates to similar performance with both \method and \pangu. With the exploration corpus, however, coverage of test patterns drops to nearly 70\%, resulting in a large 30.93 point drop using \pangu~and only 6.79 with \method, highlighting the strong compositional generalizability of our method.

\paragraph{\method with exploration is competitive with supervised ICL.} We observe that \method~+~\exploration~is able to nearly match \method~+~\trainset~(row 4 and 7 in Table~\ref{tab:kgqa_grailqa}) on GrailQA. Notably, we find that unsupervised \method is, in fact, able to \textit{outperform} supervised \pangu~when the underlying base model is held constant (MPT-7B). On MetaQA, the gap between \method~+~\exploration~and supervised ICL is a larger 6.79 F1, which can be explained by the formulaic nature of questions in MetaQA, resulting in all patterns being covered by the training set (see Table~\ref{tab:exploration_coverage}). Overall, our results demonstrate that exploration is a viable means to provide unsupervised grounding for reasoning.

\paragraph{\method with exploration leads to more consistent performance across generalization splits versus supervised methods.} In Table~\ref{tab:kgqa_grailqa}, we find that \method~+~\exploration~demonstrates low variance (2.35 versus 7.09 standard deviation using \exploration~and \trainset, respectively) in performance across generalization splits while methods using training data show fluctuations (drops) in performance on both compositional and zero-shot splits. We argue that the unsupervised nature of exploration allows \method to discover reasoning patterns without additional bias introduced by a training distribution, thus allowing it to generalize well. 

\paragraph{\method improves with model scale.} To evaluate potential gains with \method by improving the underlying LLM, we compare KGQA performance using MPT-7B versus MPT-30B and GPT-3.5, a state-of-the-art instruction-tuned LLM from OpenAI. Due to a limited budget of \$100, we sample a small set of 300 questions from the GrailQA dev set and evaluate \method~+~\exploration. Table~\ref{tab:kgqa_grailqa_gpt} shows that improving the base model indeed leads to consistent gains in KGQA performance, with MPT-30B and GPT-3.5 showing improvements of 2.79 and 8.37 F1, respectively. \method~+~GPT-3.5 additionally demonstrates more consistent performance across generalization splits as compared to \pangu-FT (state-of-the-art) and, notably, \textit{outperforms} it on zero-shot queries by 7.08 F1.
\label{subsec:results:scaling}

\begin{table}[htb]
    \small
    \centering
    \begin{tabular}{l c c c}
    \toprule
        \bf Method & \bf Overall & \bf I.I.D. & \bf Zero-shot \\
        \midrule
        Zero-shot & 15.92 & 13.75 & 22.42 \\
        \bf \method~+~\exploration & \textbf{62.25}~($\Delta$+46.33) & \textbf{63.85} & \textbf{57.44} \\
    \bottomrule
    \end{tabular}
    \caption{\textbf{KGQA Results on MatKG.} F1-scores for \method with 9,445 explored programs on a test set of 100 questions (75/25 i.i.d./zero-shot) compared to a zero-shot baseline using MPT-7B.}
    \label{tab:kgqa_matkg}
\end{table}

\paragraph{Case Study: Materials Science KG.} 
We, next, evaluate the ability of \method to work in arbitrary, specialized domains by creating a natural language interface for an unseen KG from materials science using \textbf{MatKG}~\citep{venugopal2022the}\footnote{To the best of our knowledge, this KG is not part of the pre-training corpus for the MPT family of models.}.
Since the graph is not accompanied by a set of natural language questions, we randomly sample 100 programs up to 3-hops with unique query patterns and manually annotate them to construct a test set (see~\ref{sec:appendix_matkg}). As shown in Table~\ref{tab:kgqa_matkg}, \method~+~\exploration~(with $|\exploration|\approx$~10K) results in a large 46.33 F1 gain over zero-shot reasoning that uses no exploration.

\section{Analyses}
\label{sec:analyses}
In this section, we present detailed analyses of \method. We first study the efficacy of graph exploration and its effect on downstream QA performance. We then conduct several ablations to validate the design choices made in \method---inverse-consistency and L2M for question generation, inverse-consistency during reasoning, prompting with KG schema descriptions, as well as candidate pruning. 

\subsection{KG and Query Coverage with Exploration}
\label{subsubsec:analysis-kg_dataset_coverage}
\paragraph{Exploration statistics.} Table~\ref{tab:exploration_stats} shows the results of unsupervised KG exploration on Freebase (Commons) as well as MoviesKG, including the distribution of programs of different complexity as well as the wall-clock time taken for the procedure. While program generation is inexpensive, the cost of question generation restricts the number of programs we can explore. We stop at 10K to meet our stated goal of readying a QA system within a day.
\begin{table}[h!]
    \small
    \centering
    \begin{tabular}{l r r }
    \toprule
        \textit{(budget of 10k programs)} & \bf Freebase & \bf MoviesKG \\
        \midrule
        \bf Programs & 10,000 & 10,000 \\
        \hspace{0.6cm} 1-hop & 6,933 & 222 \\
        \hspace{0.6cm} 2-hop & 2,589 & 1,779 \\
        \hspace{0.6cm} 3-hop & 426 & 4,290 \\
        \hspace{0.6cm} 4-hop & 52 & 3,709 \\
        \bf Relations & 4,178 & 18 \\
        \bf Classes & 1,681 & 7 \\
        \bf Patterns & 7,193 & 3,658 \\
        \bf Sub-expressions & 7,741 & 71 \\
        \bf Time && \\
        \hspace{0.6cm} Exploration (mins) & 46.5 & 24.4  \\
        \hspace{0.6cm} Query Generation (hours) & 10.4 & 24.0  \\
    \bottomrule
    \end{tabular}
    \caption{\textbf{Exploration Statistics} on Freebase and MoviesKG for a budget of 10K programs (capped at 5 programs per query pattern) using 
    3 Amazon EC2 p3dn.24xlarge machines.
    (\textit{Note:} relation counts listed also include reverse relations.)}
    \label{tab:exploration_stats}
\end{table}

\begin{table}[]
    \small
    \centering
    \begin{tabular}{l cc cc}
    \toprule
        & \multicolumn{2}{c}{\bf GrailQA} & \multicolumn{2}{c}{\bf MetaQA} \\
        \cmidrule(lr){2-5}
        \textit{(in dev set)} & \bf \fulltrainset & \exploration & \bf \fulltrainset & \exploration \\
        \midrule
        \bf Relations & 82.49 & 76.89 & 100.00 &  100.00 \\
        \bf Classes & 85.43 & 91.56 & 100.00 & 100.00 \\
        \bf Patterns & 70.93 & \underline{13.94} & 100.00 & \underline{69.39} \\
        \bf Sub-expressions & 79.24 & \underline{49.43} & 100.00 & 100.00 \\
    \bottomrule
    \end{tabular}
    \caption{\textbf{Distribution Coverage with Exploration (\exploration)} versus the full training data (\fulltrainset) for queries in the development sets. On MetaQA, \exploration~provides high coverage (though nearly 30 points below \fulltrainset~on query pattern coverage) due to the small size of MoviesKG. On GrailQA, with the larger Freebase KG, \exploration~shows a huge 56.99 points drop in query pattern coverage as well as a 29.81 drop for sub-expressions, leading to several queries being zero-shot versus when using the training data.}
    \label{tab:exploration_coverage}
\end{table}

\paragraph{Distribution coverage.} To effectively ground reasoning in \method, exploration must be able to provide sufficient coverage over the queries being evaluated. We analyze how well our random exploration strategy with a budget of 10K performs compared to a curated training set in providing coverage over the evaluation distribution. Table~\ref{tab:exploration_coverage} shows our results for coverage over relations, classes, program patterns, and sub-expressions (e.g. ``\texttt{(COUNT \#var)}'', ``\texttt{(ARGMIN type.datetime \#var)}'') found in the gold logical programs from the dev sets of GrailQA and MetaQA. 

On MetaQA, we find that while exploration \textit{can} find all schema items and sub-expressions, it misses nearly 30\% of program patterns in the test distribution while the training set has perfect coverage. On GrailQA, both sub-expression and pattern coverage are much lower than training, with \exploration~observing 5x fewer test patterns and 1.6x fewer test sub-expressions than the training data. These gaps explain the difference in performance between supervised methods and \method~+~\exploration, which is completely zero-shot (Table~\ref{tab:kgqa_grailqa} and Table~\ref{tab:kgqa_metaqa}). This gap also highlights a future direction for improving \method by incorporating more guidance into exploration that goes beyond diversity alone.

\begin{figure}
    \hspace*{-0.5cm}
    \centering
    \includegraphics[width=\linewidth]{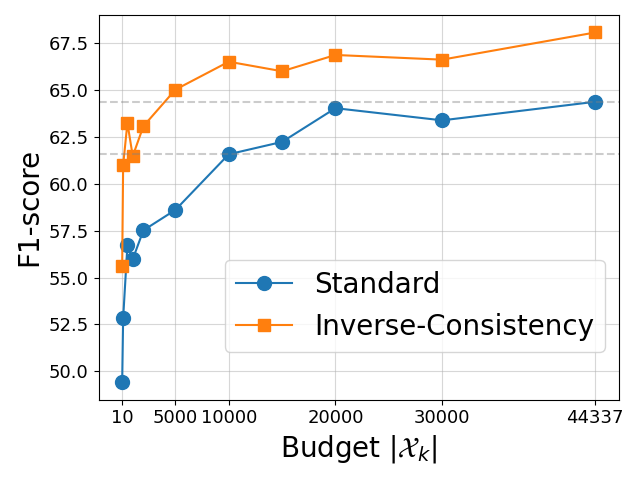}
    \caption{\textbf{Accuracy v/s Exploration Budget.} F1-scores with \method~+~$\exploration_k$~
    using MPT-7B. 
    \method shows consistent gains with increasing exploration budget, notably showing a positive slope even at the maximum budget, indicating room for further improvement. Further, inverse-consistency candidate re-ranking improves performance at all budget levels and outperforms standard predictions at $\exploration_k=$~10K with only 500 programs (20x reduction) and $\exploration_k=$~44K with only 5K programs (9x reduction).}
    \label{fig:budget_kgqa}
\end{figure}

\subsection{QA Accuracy v/s Exploration Budget}
\label{subsubsec:analysis-budge_qa_perf}
As shown in Table~\ref{tab:exploration_coverage}, real-world KGs, such as Freebase, are intractable to exhaustively explore resulting in only approximate coverage. Here, we evaluate the budget-accuracy trade-off of \method, i.e. how the \textit{amount} of exploration affects downstream QA performance. For this analysis, we randomly sub-sample multiple sets $\exploration_k$~of varying sizes $k$ from \exploration, which we then use to answer questions over a sub-sampled set of 3,000 questions (1k from each split) from the GrailQA dev set. In Fig~\ref{fig:budget_kgqa}, we plot F1-scores for \method~+~$\exploration_k$. \method shows steady improvements with more exploration, notably showing a positive slope even at 44K programs (our maximum due to budget constraints). \\

\noindent \textbf{Inverse-consistency.} Additionally, Fig.~\ref{fig:budget_kgqa} shows that re-ranking improves performance at all budget levels. Notably, re-ranking recovers (and exceeds) the performance of standard predictions at the maximum budget with only a small set of 500 programs, i.e. a 20x reduction in exploration cost, which translates to a wall-clock setup time of only 1.6 hours (versus 1.3 \textit{days} for 10K programs). Additionally, performance at the maximum budget of 44K programs can be matched using only 5K programs with inverse-consistency (9x reduction).

\subsection{Ablations}
\label{sec:ablations}
\ulb{(a)} In Appendix~\ref{subsubsec:analysis-inv_consistency_q_reranking} and Appendix~\ref{subsubsec:analysis-l2m}, we verify the efficacy of inverse-consistency re-ranking and L2M for question generation. On human evaluations, we find that inverse-consistency provides a large 22.5 point gain in semantic accuracy and L2M results in a gain of 17.5 points. Additionally, we include an ablation in Appendix~\ref{subsubsec:analysis-inv_consistence_ans_reranking} to show that inverse-consistency also improves reasoning accuracy ($\Delta$+4.94 and $\Delta$+0.83 F1 on GrailQA and MetaQA, respectively).
\ulb{(b)} In Appendix~\ref{app: schema_supervision}, we provide an ablation to verify the beneficial effect of providing natural language schema descriptions to the LLM for question generation. 
\ulb{(c)} Finally, in Appendix~\ref{subsubsec:analysis-cand_pruning}, we analyze the effect of candidate pruning during reasoning and find that our most aggressive setting ($k=$~10) not only reduces inference cost/query to 13s (8x $\downarrow$ v/s no pruning) but also results in greater accuracy ($\Delta$+2.5 F1).

\section{Related Work}
\label{sec:relatedwork}
\paragraph{KGQA Generalization.} KGQA beyond i.i.d. samples has seen 
progress
both in terms of new benchmarks \citep{gu2021beyond, dutt2023designing} as well as methods \citep{yu2023decaf, shu-etal-2022-tiara, ye-etal-2022-rng, gu-su-2022-arcaneqa}. Recently, works have also investigated generalization to unseen KGs \citep{dutt-etal-2022-perkgqa, gao2023double}. However, these methods all assume access to \emph{some} curated training data, which is completely unavailable in our 
unsupervised setting. 
We also highlight Bio-SODA \citep{sima-etal-biosoda-2021}, which shares our unsupervised setting. Their approach uses string similarity to match query tokens with KG schema items, rank them using a PageRank-based importance measure, construct a query graph using Steiner trees, and finally convert the graphs into SPARQL queries. However, 
this method is unable to handle complex queries --- aggregations, superlatives, comparatives, conjunctions, amongst others. In concurrent work, \citet{li2023flexkbqa} propose a method to train KGQA models from synthetic data using LLMs. Unlike \method, however, their work utilizes unlabeled queries from the train set as weak supervision and is, thus, not fully unsupervised.
Beyond structured queries, our work is also related to PAQ \citep{lewis-etal-2021-paq}, which over-generates questions over Wikipedia but, crucially, returns only a cached response at test time instead of reasoning as in \method.

\paragraph{KGQA with ICL.} Many recent works have attempted to unify LLMs and knowledge graphs \citep{tian2023graph, tan2023make, li-etal-2023-shot}. However, most prior works require a training corpus to retrieve in-context demonstrations, which is unavailable in our setting. A prior work that does operate in a completely zero-shot setting is 
~\citet{baek2023knowledge}, where triples are retrieved from the KG to generate the final answer. However, this method does not provide the answer text alone due to a generative strategy\footnote{They use a ``generative accuracy'' metric, which considers a prediction correct if the tokens of an answer entity are found anywhere within the generated text.} making it largely incomparable with \method.

\paragraph{Grounded Multi-Step Reasoning.} Bottom-up parsing iteratively builds a solution for complex problems 
in several prior works in semantic parsing \citep{rubin-berant-2021-smbop, gu-su-2022-arcaneqa, ye-etal-2022-rng, gu-etal-2023-dont}. \method further grounds each step of bottom-up parsing to the KG using a case-based reasoning (CBR) approach, which has widely been applied in various tasks, such as link prediction \citep{das2022knowledge}, semantic parsing \citep{das-etal-2021-case, awasthi-etal-struct-2023}, and reading comprehension \citep{thai2023machine}. \footnote{Please refer to Appendix~\ref{sec:appendix_related_work} for further related work.}

\section{Conclusion}
\label{sec:conclusion}
We introduce \textbf{\method}---a universal KGQA system to work with \emph{any} target KG and without \emph{any} human-annotated training data.
\method mimics curiosity-driven learning in humans by first exploring the unseen KG, followed by using the acquired knowledge to answer questions. 
Our method combines LLMs with
graph traversal to explore the KG
and then
reason over the explored paths to answer arbitrary user queries over the graph.
We further introduce techniques to improve zero-shot performance with LLMs, including an inverse-consistency re-ranking method.
On two popular datasets and KGs, we demonstrate the efficacy of \method and present detailed analyses of the several design choices.

\section*{Acknowledgements}
We thank Sandesh Swamy, Vinayshekhar Bannihatti Kumar, Sailik Sengupta, Sam Davidson, Bryan Li, and other members of AWS AI Labs for their helpful comments and suggestions throughout the project. 
We thank Nishant Yadav, Andrew Drozdov, Andrew McCallum, and other members of IESL for helpful discussions and advice.
We thank Vineeth Venugopal and Elsa Olivetti (MIT) for useful feedback on using MatKG.
Lastly, we are also grateful to Yu Gu and Yu Su for providing excellent documentation and support for their KGQA research.

\section*{Limitations}
\label{sec:limitations}

\paragraph{Hallucinations.} Despite efforts to provide grounded LLM generations, \method is susceptible to hallucinations at times. For instance, during question generation, models may generate semantically inaccurate queries for their corresponding programs. While we do observe robustness to noise during exploration, inaccuracies at scale may be crippling to retrieval-augmented reasoning, which is reliant on coherent exemplars for candidate scoring. 
Future directions may explore using models pre-trained for KGQA or even KG-specific parameter tuning to mitigate these behaviors.

\paragraph{Latency.} While we reduce latency by 8x compared to a naive implementation by introducing candidate pruning, our iterative ``System 2'' reasoning may not satisfy stringent response time requirements, which are better served by single-shot inference. Caching does address this limitation to an extent, but future work may explore how programs can be synthesized more efficiently for complex, multi-hop queries.

\paragraph{Zero annotations.} The primary goal of \method is to provide a query interface without any human intervention. However, as a prerequisite, we assume the availability of a schema enumerating the classes and relations present in the KG along with their natural language descriptions. Our assumption is based on the common availability of such a file accompanying most real-world KGs. In the absence of this data, \method, thus, currently requires human annotations. Further leaning on the broad-spectrum generalization abilities of LLMs, future work may explore automatically generating such schema descriptions.

\section*{Broader Impact}
\label{sec:ethics}

Our method has the potential to improve information access in several domains that contain structured information but lack the human expertise or resources to construct complex query interfaces, improving the availability of information in previously opaque settings. 

However, we caution that non-deterministic systems, such as those using LLMs, should be deployed in real-world settings with utmost care and proper human oversight. In particular, it may not always be apparent as to the nature of the data that LLMs are pre-trained on, which has the potential of perpetuating factual inaccuracies and biases prevalent in corpora collected from the internet. Indeed, \method is also not immune to these pathologies and future research should study methods to detect and prevent such behaviors.

\bibliography{anthology,custom}

\clearpage
\appendix

\section*{Appendices}
We provide several supplementary details of our work and organize them as follows:
\begin{itemize}
    \item Appendix~\ref{sec:analyses}: \textbf{Analyses and Ablations}
    \item Appendix~\ref{sec:appendix_matkg}: \textbf{MatKG Dataset}
    \item Appendix~\ref{sec:appendix}: \textbf{Implementation Details}
    \item Appendix~\ref{sec:appendix_related_work}: \textbf{Related Work}
    \item Appendix~\ref{app: lm_prompts}: \textbf{Language Model Prompts}
    \item Appendix~\ref{app: qualitative_examples}: \textbf{Qualitative Examples}
\end{itemize}

\section{Appendix: Analyses and Ablations}
\label{sec:analyses}
In this section, we present a detailed analysis of the design choices made in \method and how they affect downstream QA performance.

\subsection{Inverse-Consistency for Question Generation}
\label{subsubsec:analysis-inv_consistency_q_reranking}
We evaluate the effect of inverse-consistency re-ranking on the quality of question generation. Table~\ref{tab:qgen_inv_consistency} shows a comparison between the top-1 generation from a standard beam-search procedure versus the inverse-consistency re-ranked output on 3,000 randomly sampled questions from the GrailQA dev set. We use three automatic generation metrics -- ROUGE-1~\citep{lin-2004-rouge}, BLEU~\citep{papineni-etal-2002-bleu}, and BERTscore~\citep{Zhang2020BERTScore} -- computed with respect to the human-annotated gold references in the dataset. Our results show that inverse-consistency indeed improves generation quality, as measured on all metrics. We further inspect 40 randomly sampled questions for semantic accuracy using both methods, and find inverse-consistency generates accurate output for 70\% of questions, 22.5 points more than standard beam-search.
\begin{table}[htb]
    \small
    \centering
    \begin{tabular}{l c c }
    \toprule
        \bf Metrics & \bf Standard & \bf Inverse-Consistency \\
        \midrule
        ROUGE-1 & 48.17 & \textbf{52.81}~($\Delta$+4.64) \\
        BLEU & 31.54 & \textbf{38.63}~($\Delta$+7.09) \\
        BERTscore & 87.17 & \textbf{88.33}~($\Delta$+1.16) \\
        Human Evaluation & 47.50 & \textbf{70.00}~($\Delta$+22.50) \\
    \bottomrule
    \end{tabular}
    \caption{\textbf{Inverse-Consistency for Question Generation.} Generation quality with inverse-consistency re-ranking compared with standard top-1 predictions from beam search using MPT-7B. Inverse-consistency improves generation quality as measured on both automatic and human evaluation metrics.}
    \label{tab:qgen_inv_consistency}
\end{table}

\begin{table}[htb]
    \small
    \centering
    \begin{tabular}{l  c c}
    \toprule
        \textbf{Model} & \textbf{Standard} & \textbf{Least-to-Most} \\
        \midrule
         MPT-7B & 55.0 & 70.0 \\ 
         MPT-30B & 60.0 & 80.0 \\ 
         \midrule
         \textbf{Mean} & 57.5 & \textbf{75.0} ($\Delta$+17.5) \\
    \bottomrule
    \end{tabular}
    \caption{\textbf{L2M Question Generation.} Human-evaluated semantic accuracy of question generation using L2M prompting versus standard single-shot generation over a random sample of 40 questions from the GrailQA dev set. L2M prompting improves accuracy of generated questions by a significant 17.5 points.}
    \label{tab:qgen_l2m}
\end{table}

\subsection{L2M for Question Generation}
\label{subsubsec:analysis-l2m}
Here, we analyze the effect of L2M-prompting for question generation compared with standard, single-shot prompting. To conduct this analysis, we annotate a set of 40 questions and verify the semantic accuracy of the generated questions with respect to the corresponding logical programs. Table~\ref{tab:qgen_l2m} shows our results, where we find that L2M prompting provides an 18.7 point improvement over standard decoding.

\subsection{Schema Supervision for Question Generation}
\label{app: schema_supervision}
\begin{table}[htb]
    \small
    \centering
    \begin{tabular}{l c c }
    \toprule
        & \bf Standard & \bf Schema \\
        \midrule
        \bf ROUGE-1 & 51.40 & \textbf{52.81}~($\Delta$+1.41) \\
        \bf BLEU & 35.99 & \textbf{38.63}~($\Delta$+2.64) \\
        \bf BERTscore & 87.59 & \textbf{88.33}~($\Delta$+0.74) \\
    \bottomrule
    \end{tabular}
    \caption{\textbf{Schema Supervision for Question Generation.} Generation quality with schema descriptions injected into the prompt compared with standard prediction with only the query using MPT-7B over 3,000 randomly sampled questions from the GrailQA dev set.}
    \label{tab:qgen_schema_sup}
\end{table}

We evaluate the effect of providing natural language schema descriptions to the LLM during question generation. As shown in Table~\ref{tab:qgen_schema_sup}, we find that schema supervision improves generation quality as measured by each automatic metric.

\begin{table}[htb]
    \small
    \centering
    \begin{adjustwidth}{-0cm}{}
    \begin{tabular}{l c c c}
    \toprule
        $k$ & \textbf{Answer-Recall} & \textbf{Answer-F1} & \textbf{Latency (sec/q)} \\
        \midrule
         $\infty$ \,\, & \hspace{-1.15cm}\tiny (Pangu) \hspace{0.35cm} \small \textbf{100.00} & 59.70 & 110.1 \\
         50 & 98.67 & \bf 63.07 & 20.2 \\ 
         20 & 95.33 & 62.95 & 15.1 \\
         10 & 84.67 & 62.20 & \bf 13.2 \\
    \bottomrule
    \end{tabular}
    \end{adjustwidth}
    \caption{\textbf{Effect of Candidate Pruning.} Performance of \method~+~\exploration~on a sub-sampled set of 300 questions from the GrailQA dev set at different pruning thresholds $k$ for candidate set $P_t$. Answer-recall is the oracle recall of the gold program, answer-F1 measures KGQA performance, and latency is the average time per question over 300 questions. Evaluation is run with one Amazon EC2 p3dn.24xlarge machine using MPT-7B \textit{without} inverse-consistency re-ranking and \textit{without} caching. Aggressive pruning at $k=$~10 results in the most efficient reasoning with an accuracy gain of 2.5 F1 over no pruning.}
    \label{tab:reasoning_pruning}
\end{table}

\subsection{Candidate Pruning for Reasoning}
\label{subsubsec:analysis-cand_pruning}
As noted in \S{\ref{subsec:reasoning}}, we introduce candidate pruning in \method~in order to bound the latency at each reasoning step. This is in contrast to \pangu, which incurs high latency due to scoring every enumerated candidate. We analyze the effect of pruning in Table~\ref{tab:reasoning_pruning} on (1) the reachability of the gold program (answer-recall), (2) KGQA F1-scores, and (3) the latency per question\footnote{In practice, we cache responses from the SPARQL engine to improve latency over time, but turn caching off for this evaluation. Also, $k=\infty$ refers to no pruning.}.
With no pruning (Pangu), we encounter prohibitive runtimes of nearly 2 minutes per query, which is substantially reduced at $k=10$ to 13s (8x speed-up).
Surprisingly, we also find that aggressive pruning ($k=10$) results in improved reasoning accuracy (+2.5 F1 v/s at $k=
\infty$). In practice, we note that the latency of \method will continue to improve as more queries are served due to caching results from SPARQL executions.

\begin{table}[htb]
    \small
    \centering
    \begin{tabular}{l  c c}
    \toprule
        \textbf{Dataset} & \textbf{Standard} & \textbf{Inverse-Consistency} \\
        \midrule
         GrailQA & 61.58 & \textbf{66.52} ($\Delta$+4.94) \\ 
         MetaQA & 82.22 & \textbf{83.05} ($\Delta$+0.83) \\ 
    \bottomrule
    \end{tabular}
    \caption{\textbf{Candidate Re-ranking with Inverse-Consistency.} F1-scores of \method~+~\exploration~with inverse-consistency re-ranking compared to standard top-1 predictions over a sub-sampled set of 3K questions from the GrailQA dev set and the MetaQA test set. Inverse-consistency improves performance on both datasets.}
    \label{tab:reasoning_inv_con_ans}
\end{table}

\subsection{Inverse-Consistency for Candidate Re-ranking}
\label{subsubsec:analysis-inv_consistence_ans_reranking}
As described in \S{\ref{subsec:reasoning}}, we find that inverse-consistency re-ranking during reasoning helps recover from errors where exploration does not provide coverage over the test questions. Table~\ref{tab:reasoning_inv_con_ans} shows a comparison of F1 accuracy with standard scoring v/s inverse-consistency re-ranked outputs. Re-ranked programs $P_\mathrm{best}$ are computed using $\mathtt{rerank(\cdot,\cdot)}$ with $\alpha=0.5$. We find that re-ranking provides a significant gain of 4.94 F1 on GrailQA, while MetaQA performance increases by 0.83. The modest gains on MetaQA, may be attributed to higher pattern and sub-expression coverage during exploration as compared to GrailQA (Table~\ref{tab:exploration_coverage}), resulting in fewer instances where re-ranking is required.

\subsection{Inverse-Consistency v/s PMI}
\label{app: inv_pmi_comparison}
\begin{table}[htb]
    \small
    \centering
    \begin{tabular}{l c c }
    \toprule
        & \bf Inverse-Consistency & \bf PMI$_\text{DC}$ \\
        \midrule
        ROUGE-1 & \bf 52.71 & 42.97~($\Delta$-9.74) \\
        BLEU & \bf 39.94 & 23.52~($\Delta$-16.42) \\
        BERTscore & \bf 88.64 & 85.78~($\Delta$-2.86) \\
    \bottomrule
    \end{tabular}
    \caption{\textbf{Re-ranking with Inverse-Consistency v/s PMI$_\text{DC}$ for Question Generation.} Generation quality as measured using automatic metrics using MPT-7B over 100 randomly sampled questions from the GrailQA dev set.}
    \label{tab:qgen_inv_pmi}
\end{table}
\begin{table}[htb]
    \small
    \centering
    \begin{tabular}{l c c }
    \toprule
        & \bf Inverse-Consistency & \bf PMI$_\text{DC}$ \\
        \midrule
        F1-score & \textbf{66.52} & 65.02~($\Delta$-1.5) \\
    \bottomrule
    \end{tabular}
    \caption{\textbf{Re-ranking with Inverse-Consistency v/s PMI$_\text{DC}$ for Reasoning.} F1-scores using MPT-7B over 3K randomly sampled questions from the GrailQA dev set.}
    \label{tab:reasoning_inv_pmi}
\end{table}

\citet{holtzman2021surface} propose the domain-conditional pointwise mutual information (PMI$_\text{DC}$) scoring function, i.e. $\log \Pr(y | x)~/~\Pr(y | x_\text{domain})$ to address the ``surface form competition'' hypothesis, which aims to explain miscalibrated outputs from LLMs, resulting in low accuracy in zero-shot settings. While our inverse-consistency formulation $\log \Pr(x | y)$ should, in theory, provide the same ordering as PMI$_\text{DC}$, we evaluate how these methods compare as re-ranking techniques in practice. We run evaluations on sub-sampled examples from the GrailQA dev set for both question generation (Table~\ref{tab:qgen_inv_pmi}) and candidate re-ranking during reasoning (Table~\ref{tab:reasoning_inv_pmi}). For question generation, we set $x_\text{domain}$ to ``\lstinline{### English Question:\n}'' and for reasoning, we set $x_\text{domain}$ to ``\lstinline{### Logical Form:\n}''. We find that in practice the methods exhibit different behaviors, with inverse-consistency outperforming PMI$_\text{DC}$ on both question generation and reasoning. A possible explanation for this variation is LLM sensitivity to the choice of prompt constructions to calculate the terms in the re-ranked expressions.

\section{Appendix: MatKG Dataset}
\label{sec:appendix_matkg}

\paragraph{Annotation Procedure.} To evaluate \method using MatKG, we annotate a set of 100 programs with natural language questions using 2 researchers from our team. In particular, we take our set of 10K explored programs and randomly sample 100 programs such that 75 are i.i.d. for the exploration set, while 25 are o.o.d. or unseen. We then randomly split the 100 questions into two sets and iteratively provide each annotator the sampled program text, natural language descriptions for the relations, and natural language descriptions for the classes in the program. The annotator is then prompted to enter a natural language question based on this information. We release our annotated dataset for reproducibility and future research under the MIT License as part of our code repository.

\paragraph{Annotation Examples.} We provide a few examples from the annotated test set:
\begin{lstlisting}
Program: (AND material (AND (JOIN material.descriptor \"Bars\") (JOIN (R synthesis_method.material) \"Ccs\")))
Query: which materials have been synthesized using ccs and can be described as bars?

Program: (COUNT (AND descriptor (AND (JOIN (R property.descriptor) \"Free Energy Diagram\") (JOIN (R characterization_method.descriptor) \"SEM Surface\"))))
Query: how many descriptors have property free energy diagram and have characterization method sem surface?

Program: (AND application (JOIN (R characterization_method.application) (JOIN (R property.characterization_method) \"Basalts\")))
Query: the characterization method of basalts has what all applications?
\end{lstlisting}

\section{Appendix: Implementation Details}
\label{sec:appendix}

\subsection{Graphs and Datasets}
\label{app: dataset_stats}
\begin{table}[htb]
    \small
    \centering
    \begin{tabular}{cc rrr }
    \toprule
    & \textbf{Split} & \bf GrailQA & \bf MetaQA & \bf MatKG\\
     \midrule
    \multirow{3}{*}{$|\mathcal{Q}|$} & Train & 44,337 & 329,282 & - \\
    & Dev & 6,763 & 39,138 & - \\
    & Test & 13,231 & 39,093 & 100 \\
    \midrule
    \multicolumn{1}{c}{$|\mathcal{R}|$} & All & 3,720 & 9 & 21 \\
    \multicolumn{1}{c}{$|\mathcal{C}|$} & All & 1,534 & 7 & 7 \\
    \multicolumn{1}{c}{$|\mathcal{E}|$} & All & 32,585 & 43,692 & 70,002 \\
    \bottomrule
    \end{tabular}
    \caption{\textbf{GrailQA, MetaQA, and MatKG Statistics.} Note that the relation counts do not include inverse relations.}
    \label{tab:dataset_stats}
\end{table}

\paragraph{Freebase} \citep{bollacker2008freebase} is a large-scale, open-domain KG containing over 100 domains, 45 million entities, and 3 billion facts. We use the \textbf{GrailQA}~\citep{gu2021beyond} dataset, which evaluates three levels of generalization---i.i.d., compositional (novel combinations of seen constructs), and zero-shot (unseen schema items)---and also features diverse questions of varying complexity (up to 4-hop) and aggregation functions (e.g. \texttt{COUNT} and comparatives). GrailQA was constructed with the help of 6,685 crowdworkers and restricts the KG to a high-quality Commons subset, which covers 86 unique domains.

\paragraph{MoviesKG} is a small-scale, domain-specific KG provided by the WikiMovies dataset~\citep{miller2016key}, containing triples that map movies to attributes such as actors, genres, and ratings. Unlike previous work, we convert the provided triples of entity labels into a structured store where entities with the same label name may be assigned different entity IDs if they represent unique concepts.\footnote{For e.g., ``Jungle Book'' may either refer to the 1967 or the 2016 movie, but would incorrectly be considered the same entity in past work. We will release a corrected set of triples and a new set of answers for MetaQA based on this change.} The accompanying dataset we use is \textbf{MetaQA}~\citep{zhang2017variational}, which consists of more than 400K multi-hop (up to 3-hop) questions. 

\paragraph{MatKG} \citep{venugopal2022the} represents the largest KG in the materials science and was automatically generated using LLMs from 4 million scientific papers resulting in 70K entities and 5.4 million unique triples, including chemistry, structure, property, application, synthesis, and characterization data as of our study (we use version 1.2 of the KG).

\subsection{Models}
\label{app: models}
\noindent \textbf{MPT-Instruct}~\citep{MosaicML2023Introducing} is a decoder-style transformer pre-trained on 1T tokens of English text and code, followed by instruction fine-tuning on the Databricks-Dolly-15k~\citep{DatabricksBlog2023DollyV2} and Anthropic Helpful and Harmless datasets~\citep{ganguli2022red}. We use the 7B model for our main experiments and also show a small-scale experiment with 30B to verify the efficacy of \method at scale. \\

\noindent \textbf{GPT-3.5}~\citep{brown2020language} is a state-of-the-art, closed-source model from OpenAI. We conduct a small-scale experiment (constrained by budget) using the \textit{text-davinci-003} variant to demonstrate the scaling behaviors of \method (\S{\ref{subsec:results:scaling}}).

\subsection{Computing Infrastructure \& Software}
\label{app: implementation_details}
For compute, we use 3 Amazon EC2 p3dn.24xlarge instances (see \url{https://aws.amazon.com/ec2/instance-types/p3/}).
Our experiments are run using PyTorch \citep{paszke2019pytorch} and utilize Huggingface for the Transformers library \citep{wolf-etal-2020-transformers} and models to access LLMs. For executing KG programs, we use \href{https://vos.openlinksw.com}{OpenLink Virtuoso SPARQL Engine} (recommended RAM is 100G). While querying the SPARQL server, we limit each request to timeout after 5s. For more details, please refer to our repository or \url{https://github.com/dki-lab/Freebase-Setup/}.

\subsection{LLM Decoding Parameters}
\label{app: llm_decoding_params}
We use the following decoding arguments with the \texttt{generate()} call of HuggingFace's \texttt{AutoModelForCausalLM}:
\begin{lstlisting}
default_decoding_args = {
    "max_new_tokens": 100,
    "do_sample": False,
    "temperature": 0.6,
    "min_length": None,
    "use_cache": True,
    "repetition_penalty": 1.,
    "length_penalty": 1.,
    "num_beams": 10,
    "num_return_sequences": 10,
    "no_repeat_ngram_size": 10,
    "renormalize_logits": True
}
\end{lstlisting}

\subsection{Reasoning Implementation}
\label{app: reasoning_details}
\paragraph{Program Expansion Heuristics.} We re-implement the Freebase expansion heuristics detailed in \citet{gu-etal-2023-dont}, to allow operating with arbitrary KGs that may then be setup with just a text file of triples.

\paragraph{Entity Linking.} For GrailQA, we utilize the entity linking results from \citet{shu2022tiara} made available by \citet{gu-etal-2023-dont}. For MetaQA, a simple string-matching approach results in perfect EL accuracy. For MatKG, we only evaluate with gold entity links, which are made available when automatically sampling programs.

\section{Appendix: Related Work}
\label{sec:appendix_related_work}
\paragraph{KGQA Generalization.}   
Another line of work investigates pipelines for constructing semantic parsers for new KGs by generating training data automatically \citep{wang-etal-2015-building, liang2016neural, su-etal-2016-generating, gu2021beyond}. Each of these methods, however, includes a human annotation step to generate the final training data whereas \method is able to operate without any supervision. 

\citet{galkin2023towards} recently introduced a foundational model to learn transferable representations for KGQA that allows them to generalize to unseen graphs without any training data. While similar in motivation to \method, they do not handle natural language queries.

\paragraph{Planning and RL.} Reasoning in \method can be seen as iteratively constructing a plan to \textit{navigate} the KG conditioned on a test query. Many prior works take a similar view and use reinforcement learning to construct path-finding algorithms for KGQA \citep{xiong-etal-2017-deeppath, das2018go}. These methods, however, were not designed to handle natural language queries. Several recent works also investigate the use of LMs as planners to navigate environments other than KGs, such as in robotics \citep{huang2022inner, huang2022language}, unstructured reasoning \citep{zaheer2022learning, yao2023react, shinn2023reflexion}, game environments \citep{wang2023voyager}, and web navigation \citep{deng2023mind2web}.

\paragraph{LM Generation Re-ranking.} Beyond LM decoding \citep{Holtzman2020The, lu-etal-2022-neurologic}, recent work has also studied how best to \emph{rank} sequences generated by LMs. For instance, \citet{krishna-etal-2022-rankgen} train an encoder model to score generations given a prefix using contrastive learning. \citet{holtzman2021surface} instead propose an alternative PMI-based scoring function to address the ``surface form competition'' hypothesis, which is related to our inverse-consistency methodology. Prior work in information retrieval \citep{sachan-etal-2022-improving, sachan-etal-2023-questions} also makes use of a similar idea to re-rank retrieved passages for QA. Our method, however, does not require any training and also demonstrates better accuracy than PMI (see Appendix~\ref{app: inv_pmi_comparison}).

\onecolumn

\section{Language Model Prompts}
\label{app: lm_prompts}
\textit{\textbf{Note:} Due to legal restrictions, we replace generation outputs from LLMs with human-written text within double-brackets (``[[...]]'') describing the output instead.}
\subsection{Question Generation: L2M}
\label{app: lm_prompts_qgen_l2m}
\textbf{Logical program}:
\begin{lstlisting}
(AND meteorology.tropical_cyclone (AND (JOIN meteorology.tropical_cyclone.category (JOIN meteorology.tropical_cyclone_category.tropical_cyclones "Tropical Storm Linda")) (JOIN meteorology.tropical_cyclone.affected_areas "turks & caicos islands")))
\end{lstlisting}
\textbf{Prompt (for the last L2M iteration):}
\begin{lstlisting}
### Instructions:
Translate the following logical form query into a natural language question in English. The generated question must have the same meaning as the logical query. The generated question must cover all and only the information present in the logical query. The generated question should use the schema which describes the entities, relations, and functions present in the logical query. Use each previous query and solution as a hint to solve the next query.

### Logical Query:                 
(AND meteorology.tropical_cyclone_category (JOIN meteorology.tropical_cyclone_category.tropical_cyclones "Tropical Storm Linda"))
### Schema:
meteorology.tropical_cyclone=tropical cyclone; meteorology.tropical_cyclone_category=tropical cyclone category; meteorology.tropical_cyclone_category.tropical_cyclones=tropical cyclones
### English Question:
[[LLM generates a question asking about the tropical cyclone category of tropical storm linda]]

### Logical Query:
(AND meteorology.tropical_cyclone (JOIN meteorology.tropical_cyclone.category (JOIN meteorology.tropical_cyclone_category.tropical_cyclones "Tropical Storm Linda")))
### Schema:
meteorology.tropical_cyclone=tropical cyclone; meteorology.tropical_cyclone_category=tropical cyclone category; meteorology.tropical_cyclone_category.tropical_cyclones=tropical cyclones; meteorology.tropical_cyclone.category=category
### English Question:
[[LLM generates a question asking about the tropical cyclone category of tropical storm linda]]

### Logical Query:
(AND meteorology.tropical_cyclone (JOIN meteorology.tropical_cyclone.affected_areas "turks & caicos islands"))
### Schema:
meteorology.tropical_cyclone=tropical cyclone; meteorology.tropical_cyclone.affected_areas=affected areas
### English Question:
[[LLM generates a question asking about the tropical cyclones that have affected the turks and caicos islands]]

### Logical Query:
(AND (JOIN meteorology.tropical_cyclone.category (JOIN meteorology.tropical_cyclone_category.tropical_cyclones "Tropical Storm Linda")) (JOIN meteorology.tropical_cyclone.affected_areas "turks & caicos islands"))
### Schema:
meteorology.tropical_cyclone=tropical cyclone; meteorology.tropical_cyclone_category=tropical cyclone category; meteorology.tropical_cyclone_category.tropical_cyclones=tropical cyclones; meteorology.tropical_cyclone.category=category; meteorology.tropical_cyclone.affected_areas=affected areas
### English Question:
[[LLM generates a question asking about both the cyclone category of tropical storm linda and the turks and caicos islands]]

### Logical Query:
(AND meteorology.tropical_cyclone (AND (JOIN meteorology.tropical_cyclone.category (JOIN meteorology.tropical_cyclone_category.tropical_cyclones "Tropical Storm Linda")) (JOIN meteorology.tropical_cyclone.affected_areas "turks & caicos islands")))
### Schema:
meteorology.tropical_cyclone=tropical cyclone; meteorology.tropical_cyclone_category=tropical cyclone category; meteorology.tropical_cyclone_category.tropical_cyclones=tropical cyclones; meteorology.tropical_cyclone.category=category; meteorology.tropical_cyclone.affected_areas=affected areas
### English Question:
\end{lstlisting}
\textbf{Generation (natural language question):}
\begin{lstlisting}
[[LLM generates the final question asking about both the cyclone category of tropical storm linda and the turks and caicos islands]]
\end{lstlisting}

\subsection{Question Generation: Inverse-Consistency Re-ranking}
\label{app: lm_prompts_inv}
\textbf{Generated question to re-rank:}
\begin{lstlisting}
[[Candidate question generated by the LLM about the cyclone category of tropical storm linda]]
\end{lstlisting}
\textbf{Prompt:}
\begin{lstlisting}
### Instructions:
Translate the following question into its semantic parse.

### Question:
[[Candidate question generated by the LLM about the cyclone category of tropical storm linda]]
### Semantic Parse:
\end{lstlisting}
\textbf{Completion to score (logical program):} 
\begin{lstlisting}
(AND meteorology.tropical_cyclone_category (JOIN meteorology.tropical_cyclone_category.tropical_cyclones "Tropical Storm Linda"))
\end{lstlisting}

\subsection{Reasoning}
\label{app: lm_prompts_reasoning}
\textbf{Test question:} 
\begin{lstlisting}
which movies was Rob Williams the writer of?
\end{lstlisting}
\textbf{Prompt:}
\begin{lstlisting}
### Instructions:
Write a logical form expression using only elements mentioned in the provided natural language question. An "R" before a relation in the logical expression may be used to indicate a reverse or inverse relation.

### Question:
[[Question generated by the LLM asking about the movies that bernard girard wrote scripts for]]
### Logical Form:
(AND movie.movie (JOIN movie.written_by "Bernard Girard"))

### Question:
[[Question generated by the LLM asking about the movies that paul solet directed]]
### Logical Form:
(AND movie.movie (JOIN movie.directed_by "Paul Solet"))

### Question:
[[Question generated by the LLM asking about the movies that amy poehler acted in and that had the same person both direct and write the movie]]
### Logical Form:
(AND movie.movie (AND (JOIN movie.starred_actors "Amy Poehler") (JOIN movie.written_by (JOIN (R movie.directed_by) movie.movie))))

### Question:
[[Question generated by the LLM asking about the movies matt reeves directed and wrote]]
### Logical Form:
(AND movie.movie (AND (JOIN movie.directed_by "Matt Reeves") (JOIN movie.written_by "Matt Reeves")))

### Question:
[[Question generated by the LLM asking about how many movies gary k wolf wrote the scripts for]]
### Logical Form:
(COUNT (AND movie.movie (JOIN movie.written_by "Gary K. Wolf")))

### Question:
which movies was Rob Williams the writer of
### Logical Form:
\end{lstlisting}

\subsection{Reasoning: Inverse-Consistency Re-ranking}
\label{app: lm_prompts_reasoning_inv}
\textbf{Candidate program to re-rank:} 
\begin{lstlisting}
(AND travel.travel_destination (JOIN (R book.book_edition.place_of_publication) (JOIN (R book.audio_book_reader.audio_books_read) m.09qbn3)))
\end{lstlisting}
\textbf{Prompt:}
\begin{lstlisting}
### Instructions:
Write a plausible question in English that can be formed from the provided logical query as a starting point. The question must contain at least all of the information present in the logical query.

### Logical Query:
(AND travel.travel_destination (JOIN (R book.book_edition.place_of_publication) (JOIN (R book.audio_book_reader.audio_books_read) m.09qbn3)))
### Plausible Question:
\end{lstlisting}
\textbf{Completion to score (test question):}
\begin{lstlisting}
what is the name of the travel destination where mircea cartarescu is published?
\end{lstlisting}

\section{Appendix: Qualitative Examples}
\label{app: qualitative_examples}
\subsection{Inverse-Consistency Re-ranking for Question Generation}
\textit{\textbf{Note:} Due to legal restrictions, we replace generation outputs from LLMs with human-written text within double-brackets (``[[...]]'') describing the output instead.}
\label{app: qualitative_examples_inv_generation}
\subsubsection{Re-ranking sequences returned by beam search}
\begin{lstlisting}
Program:
(AND religion.founding_figure (JOIN religion.founding_figure.religion_founded (JOIN religion.religion.founding_figures "st. peter")))

Standard predictions (top-5, in order of log-probability scores):
[[Question generated by the LLM asking who paul the apostle was]]
[[Question generated by the LLM asking who christianity was founded by]]
[[Question generated by the LLM asking who the founder of christianity was ]]
[[Question generated by the LLM asking about the founding figures of the religion founded by st.peter]]
[[Question generated by the LLM asking about the founding figure of the religion founded by st. peter]]

Inverse-consistency predictions (top-5, in order of inverse-task log-probability scores):
[[Question generated by the LLM asking about the founding figure of the religion founded by st. peter]]
[[Question generated by the LLM asking about the founding figures of the religion founded by st.peter]]
[[Question generated by the LLM asking who the founder of christianity]]
[[Question generated by the LLM asking who christianity was founded by]]
[[Question generated by the LLM asking who paul the apostle was]]
\end{lstlisting}

\subsubsection{Prediction Examples}
\begin{lstlisting}
Program:
(COUNT (AND biology.breed_temperament (AND (JOIN biology.breed_temperament.breeds (JOIN biology.animal_breed.place_of_origin "swiss confederation")) (JOIN biology.breed_temperament.breeds "Toy Bulldog"))))

Standard prediction:
[[Question generated by the LLM asking about the number of dog breeds native to switzerland]]

Inverse-consistency prediction:
[[Question generated by the LLM asking about the breed temperaments of breeds originated from the swiss confederation and which also are part of the toy bulldog breed]]

---

Program:
(AND medicine.medical_trial (JOIN medicine.medical_trial.treatment_being_tested "Stavudine"))

Standard prediction:
[[Question generated by the LLM asking about the treatments that are being tested in medical trials]]

Inverse-consistency prediction:
[[Question generated by the LLM asking about the medical trials in which treatment stavudine is being tested]]

---

Program:
(AND medicine.contraindication (JOIN medicine.contraindication.contraindication_for (JOIN medicine.medical_treatment.contraindications (JOIN medicine.contraindication.contraindication_for "Teriparatide"))))

Standard prediction:
[[Question generated by the LLM asking about why teriparatide is contraindicated]]

Inverse-consistency prediction:
[[Question generated by the LLM asking about what the contraindications are for teriparatide]]

---

Program:
(AND measurement_unit.volume_unit (JOIN measurement_unit.volume_unit.measurement_system (JOIN measurement_unit.measurement_system.molar_heat_capacity_units "Joule per mole per kelvin")))

Standard prediction:
[[Question generated by the LLM asking about the molar heat capacity of joule per molecule per kelvin]]

Inverse-consistency prediction:
[[Question generated by the LLM asking about the units of volume that have a molar heat capacity units of 'joules per mole per kelvin']]

\end{lstlisting}

\subsection{Inverse-Consistency Re-ranking for Reasoning}
\label{app: qualitative_examples_inv_reasoning}
\begin{lstlisting}
Test Query:
what fictional universe does the harry potter take place in?

Standard predictions (top-5, in order of log-probability scores):
(AND fictional_universe.work_of_fiction (JOIN (R fictional_universe.fictional_universe.literary_series_set_here) (JOIN (R fictional_universe.work_of_fiction.part_of_these_fictional_universes) m.078ffw)))
(AND fictional_universe.fictional_universe (JOIN fictional_universe.fictional_universe.literary_series_set_here m.078ffw))
(JOIN (R fictional_universe.work_of_fiction.part_of_these_fictional_universes) m.078ffw)
(AND fictional_universe.fictional_universe (JOIN (R book.literary_series.fictional_universe) m.078ffw))

Inverse-consistency predictions (top-5, in order of inverse-task log-probability scores):
(AND fictional_universe.fictional_universe (JOIN fictional_universe.fictional_universe.literary_series_set_here m.078ffw))
(JOIN (R fictional_universe.work_of_fiction.part_of_these_fictional_universes) m.078ffw)
(AND fictional_universe.fictional_universe (JOIN (R book.literary_series.fictional_universe) m.078ffw))
(AND fictional_universe.work_of_fiction (JOIN (R fictional_universe.fictional_universe.literary_series_set_here) (JOIN (R fictional_universe.work_of_fiction.part_of_these_fictional_universes) m.078ffw)))

---

Test Query:
the website which had the api digg api was owned by who?

Standard predictions (top-5, in order of log-probability scores):
(JOIN (R internet.api.site) m.02hz97f)
(JOIN (R internet.website.owner) (JOIN (R internet.api.site) m.02hz97f))
(JOIN (R internet.api.protocols) m.02hz97f)
(JOIN (R internet.website.owner) (JOIN internet.website.api (JOIN (R internet.api.protocols) m.02hz97f)))

Inverse-consistency predictions (top-5, in order of inverse-task log-probability scores):
(JOIN (R internet.website.owner) (JOIN (R internet.api.site) m.02hz97f))
(JOIN (R internet.website.owner) (JOIN internet.website.api (JOIN (R internet.api.protocols) m.02hz97f)))
(JOIN (R internet.api.site) m.02hz97f)
(JOIN (R internet.api.protocols) m.02hz97f)

---

Test Query:
name the measurement system that uses newton per metre as a surface tension unit.

Standard predictions (top-5, in order of log-probability scores):
(JOIN (R measurement_unit.surface_tension_unit.tension_in_newtons_per_meter) m.02sj4sk)
(JOIN measurement_unit.measurement_system.surface_tension_units m.02sj4sk)
(AND (JOIN measurement_unit.measurement_system.surface_tension_units m.02sj4sk) (JOIN (R measurement_unit.surface_tension_unit.measurement_system) m.02sj4sk))
(JOIN (R measurement_unit.surface_tension_unit.measurement_system) m.02sj4sk)


Inverse-consistency predictions (top-5, in order of inverse-task log-probability scores):
(JOIN (R measurement_unit.surface_tension_unit.measurement_system) m.02sj4sk)
(JOIN measurement_unit.measurement_system.surface_tension_units m.02sj4sk)
(AND (JOIN measurement_unit.measurement_system.surface_tension_units m.02sj4sk) (JOIN (R measurement_unit.surface_tension_unit.measurement_system) m.02sj4sk))
(JOIN (R measurement_unit.surface_tension_unit.tension_in_newtons_per_meter) m.02sj4sk)

---

Test Query:
kg/m3 is the density units for which system of measurement?

Standard predictions (top-5, in order of log-probability scores):
(AND measurement_unit.unit_of_density (JOIN measurement_unit.unit_of_density.measurement_system (JOIN measurement_unit.measurement_system.density_units m.0d1kg)))
(AND measurement_unit.unit_of_surface_density (JOIN measurement_unit.unit_of_surface_density.measurement_system (JOIN measurement_unit.measurement_system.density_units m.0d1kg)))
(JOIN measurement_unit.unit_of_density.measurement_system (JOIN measurement_unit.measurement_system.density_units m.0d1kg))
(JOIN measurement_unit.measurement_system.density_units m.0d1kg)

Inverse-consistency predictions (top-5, in order of inverse-task log-probability scores):
(JOIN measurement_unit.measurement_system.density_units m.0d1kg)
(AND measurement_unit.unit_of_density (JOIN measurement_unit.unit_of_density.measurement_system (JOIN measurement_unit.measurement_system.density_units m.0d1kg)))
(JOIN measurement_unit.unit_of_density.measurement_system (JOIN measurement_unit.measurement_system.density_units m.0d1kg))
(AND measurement_unit.unit_of_surface_density (JOIN measurement_unit.unit_of_surface_density.measurement_system (JOIN measurement_unit.measurement_system.density_units m.0d1kg)))

---

Test Query:
what is the name of the exhibition that has the same exhibition curator with y lle celf?

Standard predictions (top-5, in order of log-probability scores):
(AND exhibitions.exhibition_curator (JOIN exhibitions.exhibition_curator.exhibitions_curated m.0w031yl))
(AND exhibitions.exhibition (JOIN exhibitions.exhibition.curators (JOIN exhibitions.exhibition_curator.exhibitions_curated m.0w031yl)))
(JOIN (R exhibitions.exhibition.curators) m.0w031yl)
(JOIN exhibitions.exhibition.curators (JOIN exhibitions.exhibition_curator.exhibitions_curated m.0w031yl))

Inverse-consistency predictions (top-5, in order of inverse-task log-probability scores):
(AND exhibitions.exhibition (JOIN exhibitions.exhibition.curators (JOIN exhibitions.exhibition_curator.exhibitions_curated m.0w031yl)))
(AND exhibitions.exhibition_curator (JOIN exhibitions.exhibition_curator.exhibitions_curated m.0w031yl))
(JOIN (R exhibitions.exhibition.curators) m.0w031yl)
(JOIN exhibitions.exhibition.curators (JOIN exhibitions.exhibition_curator.exhibitions_curated m.0w031yl))
\end{lstlisting}

\end{document}